\documentclass{article}

\usepackage{times}
\usepackage{graphicx} 
\usepackage{epsfig} 
\usepackage{subfigure} 

\usepackage{amsmath}
\usepackage{amssymb}
\usepackage{multirow}

\usepackage{bm}

\usepackage{soul} 
\usepackage{color}
\definecolor{aquamarine}{rgb}{0.5, 1.0, 0.83}

\newcommand{\hlah}[1]{#1}
\newcommand{\hlahh}[1]{#1}

\newcommand{\argmax}{\mathop{\rm arg~max}\limits}
\newcommand{\argmin}{\mathop{\rm arg~min}\limits}

\newcommand{\slinks}{{\it strong links}}

\newcommand{\wlinks}{{\it weak links}}

\newcommand{\links}{W}

\usepackage{natbib}

\usepackage{algorithm}
\usepackage{algorithmic}

\usepackage{hyperref}

\usepackage[accepted]{icml2017}

\icmltitlerunning{Outlier Cluster Formation in Spectral Clustering}

\begin{document} 

\twocolumn[
\icmltitle{Outlier Cluster Formation in Spectral Clustering}



\icmlsetsymbol{equal}{}

\begin{icmlauthorlist}
\icmlauthor{Takuro Ina}{1}
\icmlauthor{Atsushi Hashimoto}{2}
\icmlauthor{Masaaki Iiyama}{3}
\icmlauthor{Hidekazu Kasahara}{3}
\icmlauthor{Mikihiko Mori}{3}
\icmlauthor{Michihiko Minoh}{3} 
\end{icmlauthorlist}

\icmlaffiliation{1}{Graduate School of Informatics, Kyoto University, Kyoto, Japan}
\icmlaffiliation{2}{Graduate School of Education, Kyoto University, Kyoto, Japan}
\icmlaffiliation{3}{Academic Center for Computing and Media Studies, Kyoto University, Kyoto, Japan}
$^1$Graduate School of Informatics, Kyoto University, Kyoto, Japan \\
$^2$Graduate School of Education, Kyoto University, Kyoto, Japan \\
$^3$Academic Center for Computing and Media Studies, Kyoto University, Kyoto, Japan

\icmlcorrespondingauthor{Takuro Ina}{c.vvvvv@googol.com}
\icmlcorrespondingauthor{Eee Pppp}{ep@eden.co.uk}

\icmlkeywords{boring formatting information, machine learning, ICML}

\vskip 0.3in
]




\begin{abstract}
Outlier detection and cluster number estimation is an important issue for clustering real data. This paper focuses on spectral clustering, a time-tested clustering method, and reveals its important properties related to outliers. The highlights of this paper are the following two mathematical observations: first, spectral clustering's intrinsic property of an outlier cluster formation, and second, the singularity of an outlier cluster with a valid cluster number. Based on these observations, we designed a function that evaluates clustering and outlier detection results. In experiments, we prepared two scenarios, face clustering in photo album and person re-identification in a camera network. We confirmed that the proposed method detects outliers and estimates the number of clusters properly in both problems. Our method outperforms state-of-the-art methods in both the 128-dimensional sparse space for face clustering and the 4,096-dimensional non-sparse space for person re-identification.

%
%
\end{abstract}

\section{Introduction}
\subsection{Background}
Currently, many sensors are distributed in the society, and they generate a large amount of data every day. Because data collected through those sensors always contain outliers, it is important for clustering methods to detect data consensuses and isolated outliers simultaneously in a clustering task. DBSCAN \cite{ester1996density}, the winner of the 2014 SIGKDD Test of Time Award, is a reliable solution for this problem; however, it does not promise good results with high-dimensional data.
Because of the recent progress in deep learning technologies, unsupervised treatments of high-dimensional data are rapidly increasing in value.


In this paper, we revisit a well-known clustering method, spectral clustering \cite{Andrew2002,Stella2003}, and reveals its property of outlier cluster formation with high-dimensional data.
Our method was inspired originally by the problem of isolated clique enumeration in graph theory \cite{ito2005linear,huffner2009isolation,ito2009enumeration}.
Enumerating isolated cliques (or isolated pseudo cliques) extracts sets of densely connected vertices as communities in a graph, leaving other vertices as outliers. 
In a clique enumeration problem, a graph is not weighted, i.e., the graph $G(V,E)$ is defined as a graph matrix $E=\{ e_{ij} | e_{ij} \in \{0,1\}, 0\leq i,j < M \} $, for a given number of vertices $M=|V|$.
This matrix can be considered as a binarization of an affinity matrix $W=\{w_{ij} | w_{ij}\in \mathbb{R}^+, 0\leq i,j < M\}$, a common form of pairwise similarity representation in spectral clustering and other clustering methods.
Our method is derived from an attempt to extend the idea of isolated pseudo clique enumeration into the form of an affinity matrix.

We focused on the fact that the objective functions of spectral clustering and isolated pseudo clique enumeration are in good correspondence.
Based on this inspiration, we found a condition under which spectral clustering sets outliers apart from inliers (i.e., yielding several inlier clusters and one outlier cluster).
More precisely, the condition is the high dimensionality of data distribution and the validity of given numbers of clusters.
This paper mathematically and experimentally reveals the validity of the condition, and, based on the condition, we develop an evaluation function that validates results of spectral clustering. By selecting the best spectral clustering results among those obtained using different number of clusters, we identify the number of clusters automatically.


\subsection{Related Work}
High-dimensional data clustering is challenging because of the so-called ``curse of dimensionality"; Euclidean distance- and geometric-density-based approaches lose sensitivity in a high-dimensional space. 
One known strategy for high-dimensional data clustering is data projection to low-dimensional subspaces.
In many cases, high-dimensional data contain dimensions that are noisy and redundant.
Such data can be clustered by finding subspaces that data lie in, rather than by using a method that finds a geometric consensus \cite{parsons2004subspace}. 
These approaches are sometimes referred to as high-dimensional data clustering.
They assume that data has a suitable subspace in which original information is well retained.
In general, the lower a subspace is, the more information is lost. 
Hence, even for such subspace-based approaches, a clustering method that can treat high-dimensional data is useful.
Note that there are numerous subspace clustering methods, and it is difficult to cover all of them in this paper. There are good surveys \cite{parsons2004subspace,assent2012clustering} for readers interested in subspace clustering.

Graph-based approaches provide another solution. Density on a graph is defined by a set of pairwise relationships and is independent of the geometry. In other words, density on a graph makes sense even when samples are distributed in a high-dimensional space.
Therefore, finding dense subgraphs in an affinity matrix $W$ is a sound approach for avoiding the ``the curse of dimensionality."
Modularity \cite{Newman2006} is a typical graph-based method for finding communities in a network. It searches a good
cluster boundary based on a subgraph's modularity; however, there is no efficient search method \cite{brandes2006maximizing} for an exact solution and an approximate solver used in \cite{Newman2006} easily falls into local maxima.
Liu proposed a shrinking and expansion algorithm (SEA) that finds dense subgraphs at a practical computational cost \cite{Hairong2013}; however, SEA does not consider isolation of each subgraph, which is important for identifying cluster boundaries and the lack of isolation evaluation results in instability, primarily in the size of subgraphs. This restricts SEA's application, for example, to moving object detection by clustering motion coherency, which yields a subgraph with a uniform within-affinity value.

Isolation of subgraphs was first investigated in regard to operations research. Ito et al. found that the isolation condition binds a clique's border efficiently and accelerates the computational efficiency of maximal clique enumeration, which is a typical NP-complete problem without the isolation condition \cite{ito2005linear,ito2009enumeration}.
H{\"u}ffner et al. discussed several definitions of isolation and its relation to clique maximality \cite{huffner2009isolation}.
Those approaches, however, do not work with a weighted graph. Affinity matrix binarization clearly loses important information for clustering and yields many overlapping cliques. Hence, these approaches are suitable only when no weights are given on graph edges, and an overlapping solution is allowed.

Spectral clustering is a typical graph-based clustering technique that gives an approximate solution for the NP-hard normalized-cut problem \cite{Andrew2002,Stella2003}.
Several computer vision applications have used spectral clustering primarily for pixel-wise clustering, in which each pixel is described in terms of low-dimensional features \cite{li2015superpixel,narayana2013coherent,Jianbo2000}.
There are also several recent studies of spectral clustering with high-dimensional data \cite{Hai2012,wu2014spectral,wang2015clustering}, but their targets are intrinsically distributed in low-dimensional sparse subspaces and the existence of outliers are not assumed.
Those efforts relied primarily on reflecting the sparsity of data into the metric on which the affinity matrix is constructed. Such subspace projection loses information in that process when the data are not sparsely distributed and contain many outliers.

From the viewpoint of outlier detection, spectral clustering is generally regarded as a method that is not robust to outliers \cite{hennig2015handbook}. This is true when the given number of clusters is equal to that of inlier clusters, which we refer to as $N_{in}$. Our discovery is that outliers form a cluster when $N_{in}+1$ is given as the number of clusters.
The outlier cluster formation is reported in only two previous papers \cite{Wang2010,li2007noise}.
Wang \& Davidson observed the outlier clusters in their experiments, but they did not know the reason for the formation. Li et al. claimed only briefly that subspace projection with an affinity matrix lead to an outlier cluster; however, their experiments were conducted with two-dimensional data and they did not verify the formation of outlier clusters but blindly trusted the formation.
In contrast to these studies, we discuss the mechanism of outlier cluster formation mathematically and, as our contribution, we propose a method that evaluates clusters obtained using spectral clustering both to detect outliers precisely and to identify the number of clusters that yields reasonable inlier clusters.

Cluster number identification has also been a central concern of clustering and have been studied for many years \cite{Christopher2006}; 
however, known standard methods, such as x-means \cite{Pelleg2000} and Dirichlet process Gaussian mixture models \cite{Blei2006} do not work in a high-dimensional space. 
Spectral gaps (SGs) provide a good heuristic for identifying the number of clusters for spectral clustering \cite{li2007noise}. Instead of SGs, Zernik proposed another criterion that analyzes eigenvectors obtained in spectral embedding \cite{zelnik2005self}; however, the outlier cluster is not considered in either criterion.
This paper provides a new criteria that is robust against outliers, on the basis of the property of outlier cluster formation.
\section{Outlier Cluster Formation by Spectral Clustering}
\label{sec:outlier_cluster_formation}
\subsection{Normalized Cuts and Spectral Clustering}
\label{sec:normalized_cuts}
Before describing the proposed method, we provide the definitions and principles of
normalized cuts and spectral clustering.
Let $G=(V, E, W)$ be a graph where $V$ is a set of vertices and $E$ is a set of edges connecting vertices. 
$W$ is an affinity matrix, a set of edge weights that
corresponds to the similarities between the vertices in each vertex pair.

First, we consider a case in which $V$ is separated  into two clusters $A$
and $B$ \cite{Jianbo2000}.
Referring to $V_{A}$ and $V_{B}$ as sets of vertices in
$A$ and $B$, respectively, the sum of edge weights between $A$ and $B$ is
obtained as
\begin{equation}
\links(V_{A},V_{B}) = \sum_{i \in V_{A}, j \in V_{B}}W_{ij}.
\label{eq:links}
\end{equation}
Using $\links(V_{A},V_{B})$, the cost of the normalized cut between $A$ and $B$, the criterion $Ncut(V_{A}, V_{B})$ for isolation between clusters, is defined as
\begin{equation}
Ncut(V_{A},V_{B}) = \frac{\links(V_{A},V_{B})}{\links(V_{A},V)} + \frac{\links(V_{A},V_{B})}{\links(V_{B},V)}.
\label{eq:Ncut}
\end{equation}
Similarly, the association within clusters $A$ and $B$, $NAssoc(V_{A},V_{B})$, a criterion
for density within clusters, is defined as:
\begin{equation}
NAssoc(V_{A},V_{B}) = \frac{\links(V_{A},V_{A})}{\links(V_{A},V)} + \frac{\links(V_{B},V_{B})}{\links(V_{B},V)}.
\label{eq:NAssoc}
\end{equation}
\par
Second, we expand the above criteria to multi-cluster cases.
Let $\Gamma_{V}^{K} = \{ V_{1}, ... , V_{K} \} $ be a separation of $V$
into $K$ clusters.
The cost of a normalized cut is rewritten as
\begin{equation}
kNcut(\Gamma_{V}^{K}) = \frac{1}{K}\sum_{l=1}^{K} \frac{\links(V_{l}, \overline{V_{l}})}{\links(V_{l}, V)},
\label{eq:kNcut}
\end{equation}
where $\overline{V_{l}}$ is the complementary set of $V_{l}$.
Association within $K$ clusters can be rewritten as:
\begin{equation}
kAssoc(\Gamma_{V}^{K}) = \frac{1}{K}\sum_{l=1}^{K} \frac{\links(V_{l}, V_{l})}{\links(V_{l}, V)}.
\label{eq:kNAssoc}
\end{equation}

Note that minimizing
$kNcut(\Gamma_{V}^{K})$ and maximizing $kAssoc(\Gamma_{V}^{K})$ are
equivalent since $kNcut(\Gamma_{V}^{K})+kAssoc(\Gamma_{V}^{K})=1$.
Minimizing $kNcut(\Gamma_{V}^{K})$ (Maximizing $kAssoc(\Gamma_{V}^{K})$) is an NP-hard problem.
Spectral clustering can solve the above minimization approximately.

\subsection{Formation of Outlier Cluster}
\label{sec:outlier_cluster}
Here, we explain the outlier cluster formation mechanism from the perspective of spectral clustering and normalized cuts.
Spectral clustering approximates normalized cuts by relaxing the problem of
assigning discrete labels to each sample to one of calculating the likelihood for
each label.
In this process, spectral clustering represents a feature of each sample as a column in a graph
Laplacian matrix, which is calculated as $L=D-W$. 
Here $D$ is the degree matrix, defined as $D_{ii}=\sum_{j=1}^{n}W_{ij}$ and $D_{ij} = 0 (i \neq j)$,
and $W$ is the affinity matrix of $G(V,E,W)$.
Spectral clustering applies an eigendecomposition to the graph Laplacian $L$ and
represents the samples by coefficients of the first $K$ eigenvectors (low-frequency component).
Small differences between individual samples, including components of $D$, are represented by eigenvectors with
smaller eigenvalues (high frequency component) and are ignored.
Similarities between outlier samples have a low variance,
particularly in high-dimensional space.
The variance will be eliminated when they projected into a low dimensional subspace.
As a result, all outlier samples will have similar coefficients of $K$ eigenvectors and thus form a cluster.

We analyze this outlier formation mechanism more precisely from the
perspective of normalized cuts.
In principle, samples in an inlier cluster must be similar.
In other words, the expectation values of the edge weights within inlier
clusters are greater than those of the edge weights from inlier to outlier clusters (and then those of edge weights within an outlier cluster).
We refer to edges within any inlier cluster as \slinks, and
other edges, i.e., edges between different clusters and those within an outlier
cluster, as \wlinks. \ 
Let $\mu_{in}$ and $\mu_{out}$ be the expectation values of weights of {\it strong} \ 
and \wlinks, respectively. 
We represent the distribution
of weights of {\it strong} \ and \wlinks \ as $\pi(\mu_{in}, \Theta_{1})$ and $\pi(\mu_{out}, \Theta_{2})$, respectively, where $\Theta_{1}$ and $\Theta_{2}$ are distribution parameters.

Here, let $V_{in}$ be an inlier cluster. Association within an inlier cluster
$V_{in}$ is rewritten with $\mu_{in}$ as
\begin{equation}
\begin{split}
\frac{\links(V_{in}, V_{in})}{\links(V_{in}, V)} =& \frac{\links(V_{in},V_{in})}{\links(V_{in}, V_{in})+ \links(V_{in}, \overline{V_{in}})} \\
\simeq & \frac{\mu_{in}|V_{in}|^{2}}{\mu_{in}|V_{in}|^{2}+\mu_{out}|V_{in}|(|V|-|V_{in}|)} \\
=& \frac{\mu_{in}|V_{in}|}{\mu_{in}|V_{in}| + \mu_{out}(|V|-|V_{in}|)},
\end{split}
\end{equation}
where $|V|$ is the number of vertices.
Similarly, association within an outlier cluster $V_{out}$ is rewritten
with $\mu_{out}$ as:
\begin{equation}
\begin{split}
\frac{\links(V_{out}, V_{out})}{\links(V_{out}, V)} \simeq&
 \frac{\mu_{out}|V_{out}|^{2}}{\mu_{out}|V_{out}||V|} = \frac{|V_{out}|}{|V|}.
\end{split}
\end{equation}
\par
Here, we consider the best assignment of an outlier $v_o$ to  maximize the association. 
When $v_{o}$ is grouped into an inlier cluster $V_{in}$, the association
in Eq. (\ref{eq:NAssoc}) is calculated as
\begin{equation}
\begin{split}
& NAssoc(\{V_{in}, v_{o}\}, V_{out}) \\
\!\!&= \frac{\links(\left\{ V_{in}, v_{o}\right\}, \left\{ V_{in}, v_{o} \right\})}{\links(\left\{ V_{in}, v_{o} \right\}, V)} + \frac{\links(V_{out}, V_{out})}{\links(V_{out}, V)} \\
\!\!&= \frac{\links(V_{in},V_{in})+2\links(v_{o}, V_{in})+ \links(v_{o},v_{o})}{\links(V_{in}, V_{in})\! +\! \links(V_{in}, \overline{V_{in}})+ \links(v_{o}, V)} 
\!+\! \frac{\links(V_{out}, V_{out})}{\links(V_{out}, V)} \\ 
\!\!&\simeq \frac{\mu_{in}|V_{in}|^{2} + 2\mu_{out}|V_{in}|}{\mu_{in}|V_{in}|^{2} + \mu_{out}|V_{in}|(|V|-|V_{in}|)+ \mu_{out}|V|} + \frac{|V_{out}|}{|V|}. 
\end{split}
\label{eq:outlier_fg}
\end{equation}

Similarly, when $v_{o}$ is grouped into outlier cluster $V_{out}$,
the association is calculated as
\begin{eqnarray}
&&\hspace{-2.5em} NAssoc(V_{in}, \{V_{out}, v_{o} \}) \nonumber \\
&\simeq&\hspace{-0.5em}
\frac{\mu_{in}|V_{in}|}{\mu_{in}|V_{in}| + \mu_{out}(|V|-|V_{in}|)} + \frac{|V_{out}+1|}{|V|}.
\label{eq:outlier_bg}
\end{eqnarray}

The difference between Eqs. (\ref{eq:outlier_fg}) and (\ref{eq:outlier_bg}) is given as follows:
\begin{eqnarray}
&&\hspace{-2.5em} NAssoc(V_{in},\{V_{out}, v_{o}\}) - NAssoc(\{V_{in}, v_{o}\}, V_{out}) \nonumber \\
&=&\hspace{-0.5em} \frac{1}{Z}(\mu_{out}|V_{in}|^{2}|V|(\mu_{in} - 2\mu_{out}) +
 2\mu_{out}^{2}|V_{in}|^{3}) + \frac{1}{|V|} \nonumber \\ 
&>&\hspace{-0.5em} \frac{1}{Z}(\mu_{out}|V_{in}|^{2}|V|(\mu_{in} - 2\mu_{out})),
\label{eq:outlier_h}
\end{eqnarray}
where
\begin{equation}
\begin{split}
&\hspace{-0.5em}Z = (\mu_{in}|V_{in}|^{2} + \mu_{out}|V_{in}|(|V|-|V_{in}|)) \\
&\hspace{-0.0em}\cdot(\mu_{in}|V_{in}|^{2} + \mu_{out}|V_{in}|(|V|-|V_{in}|) + \mu_{out}|V|).
\end{split}
\end{equation}

$v_{o}$ increases the association when grouped into an outlier cluster as long as Eq. (\ref{eq:outlier_h}) is positive. \hlah{Clearly, Eq. }(\ref{eq:outlier_h})\hlah{ is positive when}
\begin{equation}
\mu_{in} > 2 \mu_{out}.
\end{equation}
\hlah{This indicates that all normalized-cut-based clustering methods intrinsically tend to form a gap between clusters with high and low affinity expectations}.


\subsection{Experimental Confirmation}\label{sect:experiment_synth}
To verify the above property experimentally, we prepared synthetic samples
$X=\{\bm{x}_i| 0\leq i < M\}$ with cluster labels $Y=\{y_i|0\leq i <M \}$, where $\bm{x}_i \in \mathbb{R}^D$ and $0\leq y_i\leq K$ ($y=0$ for outliers, otherwise inliers). $M$ and $N_{in}$ are the numbers of samples and the inlier clusters, respectively. Note that $X$ corresponds to $V$ and $Y$ provides a separation $\Gamma_{V}^{K}$ in the framework of normalized cuts.
Let $x_d$ be the value of $\bm{x}_i$ at the $d$\textsuperscript{th} dimension.
Outliers and cluster centers are sampled from the $D$-dimensional uniform distribution ${\cal U}^D(-1,1)$, i.e., $x_d\in {\cal U}(-1,1)$.
Along with the cluster center $\bm{x}^c = (x_1^c,\ldots,x_D^c)$, $x_d$ of an inlier $\bm{x}$ is sampled from a normal distribution ${\cal N}(\mu = x_d^c,\sigma = 1.0)$. We set $K=5$ and $M=300$, with 30 samples for each cluster and 150 outliers (100\% against inliers). Affinity (or distance for DBSCAN) is calculated on the basis of cosine similarity, which is known to be an effective metric in a high-dimensional space.\footnote{We confirmed that DBSCAN using cosine distance always achieves better scores in this experimental setting than that using the Euclidean distance.}

\begin{figure}[t]
\begin{tabular}{c}
\begin{minipage}{1.0\hsize}
\hspace{-0.5em}\includegraphics[width=\textwidth]{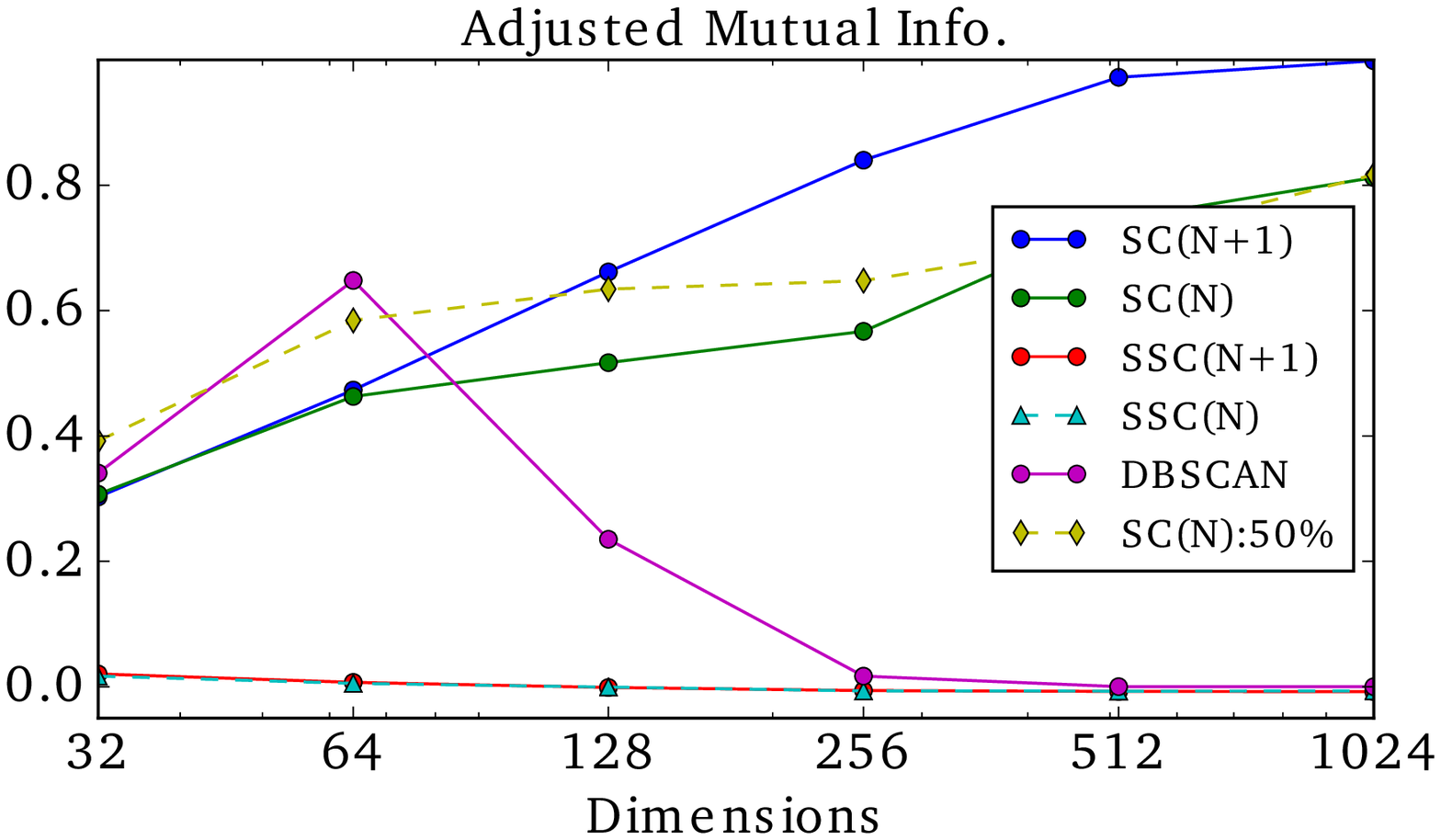}
\end{minipage}\\%
\begin{minipage}{1.0\hsize}
\hspace{-0.5em}\includegraphics[width=\textwidth]{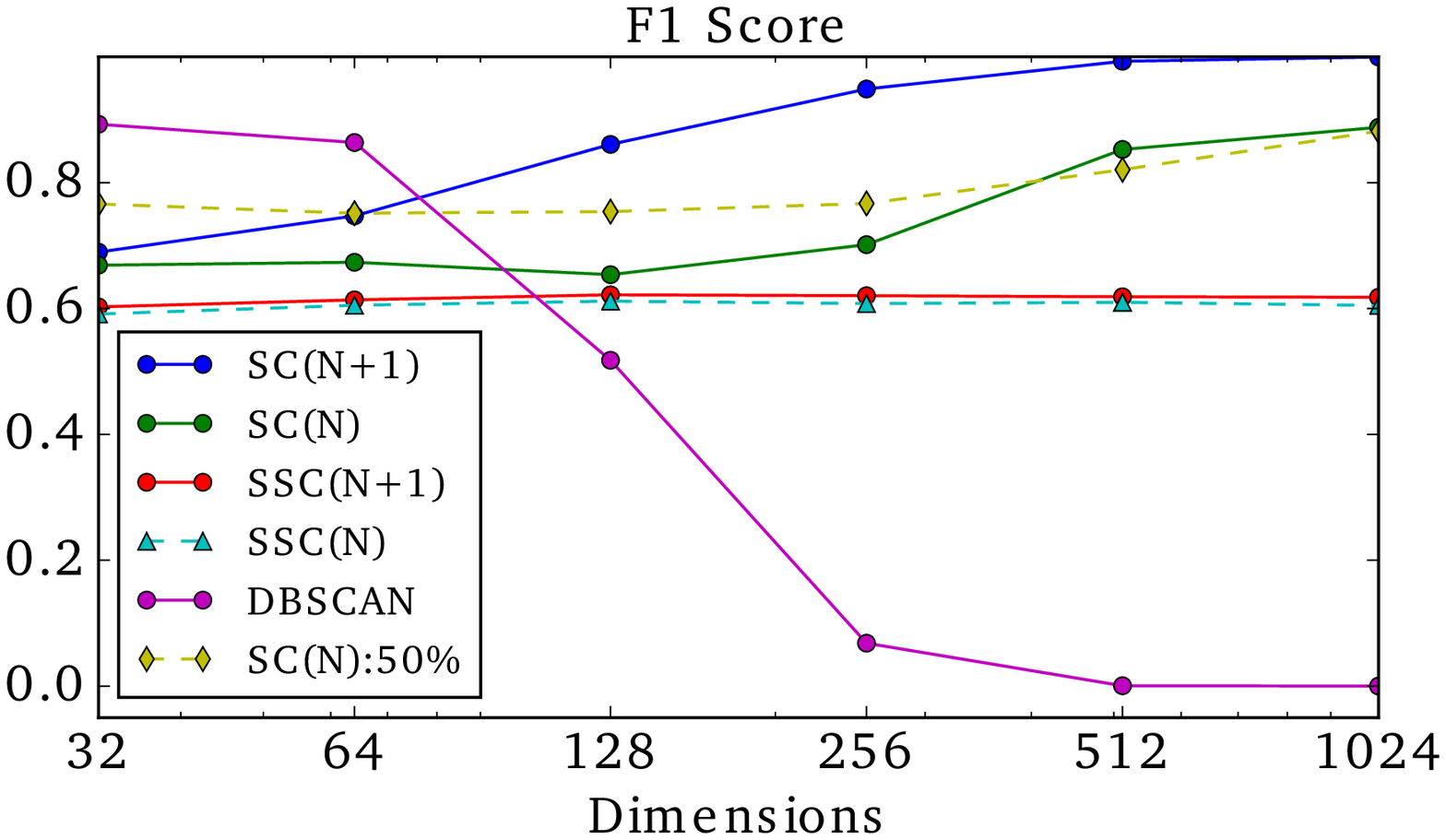}
\end{minipage}%
\end{tabular}
\caption{AMI of clustering results and F1 score of inlier/outlier classifiction. SC(N+1)/SSC(N+1) is spectral clustering and sparse spectral clustering with $K=N_{in}+1$, where $K$ is the number of clusters provided to the algorithm, and SC(N)/SSC(N) is those with $K=N_{in}$. SC(N)50\% obtained data in which the number of outliers is 50\% of that of the inliers, whereas others included 100\% }
\label{fig:outlier_removal}
\end{figure}

Figure \ref{fig:outlier_removal} shows the average scores in 100 trials.
We compared the result with DBSCAN \cite{ester1996density}, sparse spectral clustering (SSC) \cite{wu2014spectral}, and normal spectral clustering (SC).\footnote{We used scikit-learn implementation for DBSCAN and core modules of SC and SSC.} F1 scores of inlier-outlier classification for the spectral-clustering-based methods were calculated under the assumption that the sparsest cluster was the outlier cluster, whereas DBSCAN originally has the functionality of outlier detection. DBSCAN has two control parameters; we grid-searched the best parameter and used the best result in adjusted mutual information (AMI) \cite{ami}. 
Because SSC has a large computational cost, we did not grid-search in this part but set parameters using our best judgement. 

Though DBSCAN showed the best AMI score on data in a space with 64 dimensions, it does not work at all in spaces with more than 256 dimensions, whereas SC works well in such spaces. The F1 scores revealed the outlier cluster formed using spectral clustering; however, SSC does not form such a cluster, and the AMI is close to zero, indicating that the result resembles clustering by chance. From these results, we confirmed that the synthesized samples are densely distributed and no appropriate sparse representation was found.
SC achieves the best scores for ARI and F1 score in high-dimensional spaces.
Interestingly, SC(N) detects outliers more accurately when the dimension increases from 256 to 512 when the numbers of outliers and inliers are equal, though this does not occur when there are only half outliers. This can be explained in terms of the fact that $\mu_{in} > 2 \mu_{out}$ is satisfied with 512-dimensional samples in both cases, that the potential outlier association $|V_{out}|/|V|$ is $1/2$ and $1/3$ in the two cases, and that $1/2$ exceeded the cost to retain two inlier clusters as independent clusters but $1/3$ did not.
This indicates that the more data contain outliers, the more stable the outlier cluster formation is.
In contrast, SC(N$+1$) does not cause the competition of the outlier and inlier clusters because $K$ has room for the outlier cluster.
\section{Evaluating Clusters by the Contrast between Inlier and Outlier Clusters}\label{sec:evaluating_clusters}
\subsection{Singularity of Outlier Clusters}
In the previous section, we have confirmed spectral clustering's tendency of outlier cluster formation mathematically and experimentally.
In this section, we discuss the number of appearing outlier clusters $N_{out}(>0)$.
Let $N_{in}$ be locally optimal numbers of inlier cluster for the given dataset, where the local optimality is defined by the following inequalities.
\begin{eqnarray}
(K-2)kAssoc(\Gamma_{V}^{N_{in}-1})
< (K-1)kAssoc(\Gamma_{V}^{N_{in}})
\label{eq:local_optimality} \\
(K-1)kAssoc(\Gamma_{V}^{N_{in}})
> K kAssoc(\Gamma_{V}^{N_{in}+1})
\label{eq:local_optimality2}
\end{eqnarray}

In addition, let ${\hat N_{in}}$ be the estimated numbers of inlier clusters, for fixed $K$ ($K={\hat N_{in}}+N_{out}$).
We show that the case $N_{out}=1$ marks the best score in the association defined in Eq. (\ref{eq:kNAssoc}) when $K=N_{in}+1$.

First, assuming that $N_{out}=1$ (${\hat N_{in}}=N_{in}$), the association is calculated as
\vspace{-0.5em}
\begin{eqnarray}
\hspace{-1.0em} kAssoc(\Gamma_{V}^{N_{in},1}) \hspace{-1.0em}&=&\hspace{-1.0em} \frac{1}{K}
 \bigl(\sum_{l=1}^{K-1} \frac{\links(V_{l},V_{l})}{\links(V_{l},V)} + \frac{|V_K^1|}{|V|}\bigr)\nonumber \\ 
 \hspace{-1.0em}&=&\hspace{-1.0em} \frac{1}{K}\bigl((K\!-\!1)kAssoc(\Gamma_{V}^{N_{in}})\!+\! \frac{|V_K^1|}{|V|}\bigr),
 \label{eq:n_out1}
\end{eqnarray}
where $\Gamma_{V}^{{\hat N}_{in},{\hat N}_{out}}$ is the set of clusters
in the first ${\hat N}_{in}$ clusters having greater $\mu_i$ than those
from the last ${\hat N}_{out}$ clusters.

Similarly, assuming $N_{out}=2$ (${\hat N_{in}}=N_{in}-1$), the association is expressed as
\begin{equation}
\begin{split}
 &kAssoc(\Gamma_{V}^{N_{in}-1,2}) \\
 &= \frac{1}{K}
 \bigl(\sum_{l=1}^{K-2} \frac{\links(V_{l},V_{l})}{\links(V_{l},V)} +
 \frac{|V_{K-1}^2|+|V_K^2|}{|V|}\bigr)\nonumber \\
\end{split}
\end{equation}
\begin{equation}
\begin{split} 
 &= \frac{1}{K}
 \bigl((K-2)kAssoc(\Gamma_{V}^{N_{in}-1})+\frac{|V_{K-1}^2|+|V_K^2|}{|V|}\bigr).
 \label{eq:n_out2}
\end{split}
\end{equation}
The difference between Eqs. (\ref{eq:n_out1}) and (\ref{eq:n_out2}) is given by
\begin{equation}
\begin{split}
 &kAssoc(\Gamma_{V}^{N_{in},1}) - kAssoc(\Gamma_{V}^{N_{in}-1,2}) \\
 &\simeq\!\! \frac{1}{K}\bigl( (K\!-\!1)kAssoc(\Gamma_{V}^{N_{in}})\!-\! (K\!-\!2)kAssoc(\Gamma_{V}^{N_{in}-1}) \bigr) \\
 &(\because |V_K^1| \simeq |V_{K-1}^2| + |V_{K}^2|)
\end{split}
\end{equation}
Furthermore, the number of outliers that satisfy $\mu_{in}>2\mu_{out}$ is
not changed by $N_{out}$; hence, we can assume that $|V_K^1| \simeq |V_{K-1}^2| + |V_{K}^2|$.
Thus, the case of $N_{out}=2$ returns a smaller association than the case of $N_{out}=1$. 
\hlah{This is obviously the same in cases $N_{out}>1$ as long as the optimality defined in the same manner with Eq. (}\ref{eq:local_optimality}\hlah{) is satisfied.
Therefore, $N_{out}=1$ is a reasonable assumption when $K=N_{in}+1$.}

\hlah{When $K\neq N_{in}+1$, $N_{out}$ can be greater than one as long as the local optimality similar to Eq. (}\ref{eq:local_optimality2}\hlah{) is satisfied, and the singularity of the outlier cluster will be violated.
Otherwise, $K\neq N_{in}+1$ and $N_{out}=1$, and $K-1$ should be another local optimal number of inlier clusters.}


\subsection{Implementation of outlier-robust evaluation function}
The previous subsection revealed the singularity of outlier cluster when $K-1$ satisfies the local optimality Eqs. (\ref{eq:local_optimality}) and (\ref{eq:local_optimality2}).
In other words, the optimality is not satisfied when the result contains multiple outlier clusters.
Using this property, we implement a simple function that evaluates the clustering result without any ground truth.

Let $\mu_i$ be the expectation of similarity
within cluster $V_i$, calculated as follows:
\begin{eqnarray}
 \mu_i = \frac{\sum_{a,b \in V_{i}, a \neq b}W_{ab}}{2|E(V_{i})|},
 \label{eq:singularity}
\end{eqnarray}
where $|E(V_{i})|$ is the number of edges within cluster $V_{i}$.
When there is only one outlier cluster, $\mu_{out}$ is estimated as
$\min \mu_i$. 
Let $V_{i_{out}}$ be the outlier cluster estimated by $i_{out} = \argmin \mu_i$.
Then, we can estimate $\mu_{in}$ by $\min_{i\neq i_{out}}\mu_i$, where
we use $\min$ rather than the average to obtain a hard estimation of
$\mu_{in}$, confirming a large difference between $\mu_{out}$ and $\mu_{in}$ for every inlier cluster.
Using the above estimated values, the separation between inliers and
outliers can be evaluated by $\mu_{in}-\mu_{out}$.

We also evaluate the isolation between inlier clusters by
\begin{eqnarray}
 \mu_{i,j} = \frac{\sum_{a \in V_{i}, b \in
 V_{j}}W_{ab}}{|E(V_{i},V_{j})|},
\end{eqnarray}
where $|E(V_{i}, V_{j})|$ is the number of edges between $V_{i}$ and $V_{j}$.
Using these values, we evaluate singularity of outlier cluster as  follows:
\begin{eqnarray}
&&\hspace{-3.5em} f(K) = \min_{i\neq i_{out}}\delta(V_i) - \delta(V_{i_{out}}), \label{eq:fcluster}
\end{eqnarray}
where $\delta(V_i)$ is $\mu_i$ based density criteria of cluster $V_i$. We defined $\delta(V_i)$ as a normalization of function $\Delta(V_i)$ as follows:
\begin{eqnarray}
&&\hspace{-3.5em} \delta(V_i) = \frac{\Delta(V_i)}{\max \Delta(V_i)}, \Delta(V_i) = \mu_i - \frac{1}{K-1}\sum_{j\neq i}\mu_{i,j}. \label{eq:Delta}
\end{eqnarray}
Note that $\Delta(V_i)$ is inspired from criteria of isolated pseudo cliques \cite{Ito2009} that offset $\mu_i$ by out-going links to emphasize isolation of each inlier cluster. 
\hlah{$\Delta(V_i)$ has the roll of selecting the best solution from the local optima filtered by the singularity of outlier cluster in Eq. }(\ref{eq:singularity}).
For searching the best solution of the number of clusters, we simply set a search range [$K_{min}$, $K_{max}$] and select $\argmax_{K_{min}\leq K \leq K_{max}} f(K)$.



\section{Experiments}
To evaluate the proposed method, we prepared two different scenarios, clustering faces appearing in a photo-album (Section \ref{seq:faceclustering}) and person re-identification with unsupervised deep-learning-based features (Section \ref{seq:personidentification}).
The first scenario, face clustering, is a practical task to index pictures in a photo album on a SNS by people's faces without supervision. We suppose inliers to be companions of SNS users and outliers to be passersby who are photographed accidentally together with the inlier people.
The second scenario, person re-identification is a demonstration to find data consensus in a deep-learning-based high dimensional space without supervision by fine-tuning.
For unsupervised learning methods, it is an important task to analyse consensus of data observed for the first time.

\par
In the experiment, we compared the results of the proposed method to state-of-the-art clustering methods that automatically estimate the number of clusters. {\bf DBSCAN} \cite{ester1996density} is used as a non-graph-based baseline method. Its parameters are chosen in the same manner to the experiments conducted in \ref{sect:experiment_synth}, and thus the parameter setting is optimal.
As graph-based clustering approaches, both spectral-clustering-based and non-spectral-clustering-based methods are selected.
{\bf SG} is a well-known heuristic method for identifying the number of clusters with spectral clustering.
Self-tuning spectral clustering ({\bf STSC}) \cite{zelnik2005self} is a state-of-the-art spectral-clustering-based method for estimating the number of clusters.
We note that both SG and STSC do not consider the outlier cluster formation.
{\bf Modularity} \cite{Newman2006} is a non-spectral-clustering-based method that can identify cluster numbers but cannot detect outliers.
{\bf SEA} \cite{Hairong2013} is a state-of-the-art method that finds dense subgraphs in a fully connected weighted graph.
The detected dense subgraphs correspond to inlier clusters, and the number of subgraphs is identified automatically. In this sense, the method is similar to the proposed method. The essential difference between SEA and the proposed method lies in the treatment of cluster isolation. SEA considers only density in its objective function, whereas our method considers both density and isolation.

In addition to the above comparative methods, we also prepared the following reference methods to which the cluster number is given.
{\bf SC(N+1)} is spectral clustering with the given number of cluster $N_{in}+1$. Because IDC is designed to estimate $N_{in}+1$ as the true cluster number, the scores of SC(N+1) are good references of IDC's ideal scores. 
{\bf SSC(N)} \cite{wu2014spectral} is a state-of-the-art spectral-clustering-based method developed for clustering high-dimensional data.
Its parameter is set using our best judgement as shown in \ref{sect:experiment_synth}.
$K=N_{in}$ is tested for sparse spectral clustering because this setting is the original usage in \cite{wu2014spectral}.

To distinguish the proposed method from the original spectral clustering, which works with a given number of clusters, we refer to the proposed method as isolated dense clustering ({\bf IDC}).

\subsection{Face Clustering in Photo Albums}
\label{seq:faceclustering}
For quantitative evaluation, we require a number of photo albums.
We synthesized such data from the Labeled Faces in the Wild dataset \cite{LFW}, a publicly available collection of images from the web that is commonly used to evaluate face detection and face re-identification methods.
The dataset includes images of 5,749 people.

\par
We prepared photo albums with the number of the companions corresponding to inlier clusters $N_{in}$ in the range of 5-15. 
In the evaluation, we conducted 100 trials with different combinations of inliers and outliers. Because spectral clustering is known to be robust when clusters have similar sizes, we varied the size of each cluster for a fair comparison. The $i$\textsuperscript{th} inlier comprises $l_i$ samples, where the number $l_i$ is randomly determined by a normal distribution ${\mathcal N}(\mu = 30,\sigma = 10)$. $l_{out}$, the number of outlier samples, is also determined by the same distribution. The outliers are $l_{out}$ people who are not inliers (i.e., we sampled one face from each person without duplication).

\par
We calculated similarity/distance between faces based on FaceNet \cite{florian2015openface}, one of the state-of-the-art face recognition methods.
FaceNet is one of the convolutional neural networks and outputs 128-dimensional sparse features, designed such that features from the same person's face will lie within a Euclidean distance of 1.0. 
For each method, we selected the better metric of the cosine similarity ($W(i,j)=\cos(x_i-x_j)+1.0$)\footnote{We added 1.0 to $\cos(x_i-x_j)$ for the spectral embedding calculation in spectral clustering that requires similarity values in affinity matrices to be positive.} and the Euclidean metric based similarity ($W(i,j) = \exp(-||x_i-x_j||/2 \gamma^{2})$, where we fixed $\gamma=1.0$).

\par
\begin{figure}[t]
\begin{center}
\includegraphics[width=0.8\linewidth]{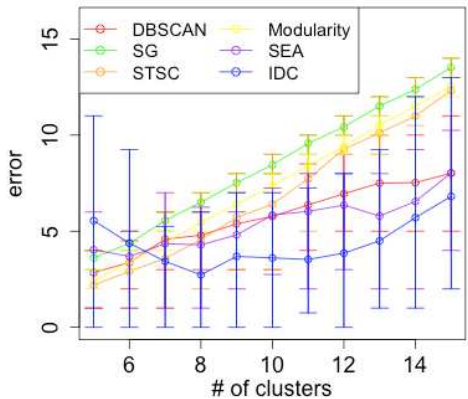}
\end{center}
\vspace{-1.5em}
\caption{Estimated number of clusters}
\vspace{-2.0em}
\label{fig:face_error}
\end{figure}

Figure \ref{fig:face_error} shows errors in the estimated number of clusters. The upper and lower bars for each method show the first and third quartiles, respectively, of the error in 100 trials. 
In the calculation, we set the lower bound of the search range for $\hat{K}$ as three by assuming multiple inlier clusters (i.e., $N_{in}=2$). The upper bound is set to 20, where this value does not affect the result seriously even when we set it to a higher value such as $M/3$ for sample number $M$.
We also set 20 to STSC as the upper bound of the search range.
IDC scored large errors for a small $N_{in}$.
This is caused because all other methods identified the number as two or three, regardless of the true cluster number.
For the same reason, the error of those methods increases linearly with $N_{in}$.
In contrast, IDC found clusters that are smaller than person-based groups. It is expected that those clusters are organized by factors such as face direction.
When $N_{in}$ increases, the variation in sample subsets grouped by such factors is also expected to increase.
In contrast, clusters grouped by the factor of person difference will not. This can explain the reason for decreasing errors in the results for IDC, which outperformed any other method for large $N_{in}$.

\begin{table*}[t]
\caption{Accuracy of face clustering by four criteria (average value in 100 trials). }
\begin{center}
\begin{tabular}{c||ccc|ccc|ccc|ccc} \hline 
score & \multicolumn{3}{c|}{Err. in \# of clusters} & \multicolumn{3}{c|}{ARI} & \multicolumn{3}{c|}{AMI} & \multicolumn{3}{c}{F1 score} \\ \hline
$N_{in}$ & 5 & 10 & 15 & 5 & 10 & 15 & 5 & 10 & 15 & 5 & 10 & 15\\ \hline
 DBSCAN & 2.84 & 5.71 & 8.04 & 0.37 & 0.33 & 0.28 & 0.53 & 0.57 & 0.56 & \textbf{0.52} & \textbf{0.36} & \textbf{0.29} \\
 SG  & 3.61 & 8.45 & 13.52 & 0.32 & 0.19 & 0.13 & 0.34 & 0.27 & 0.22 & - & - & - \\ 
 STSC & \textbf{2.19} & 6.40 & 12.32 & 0.55 & 0.37 & 0.20 & 0.56 & 0.43 & 0.28 & - & - & - \\
 Modularity & 2.41 & 7.37 & 12.53 & 0.38 & 0.21 & 0.13 & 0.40 & 0.30 & 0.23 & - & - & - \\
 SEA & 4.03 & 5.84 & 8.03 & 0.29 & 0.17 & 0.13 & 0.36 & 0.28 & 0.23 & 0.06 & 0.04 & 0.03 \\
 IDC (ours) & 5.55 & \textbf{3.61} & \textbf{6.81} & \textbf{0.60} & \textbf{0.69} & \textbf{0.52} & \textbf{0.62} & \textbf{0.72} & \textbf{0.59} & 0.36 & 0.26 & 0.23 \\
\hline
 SC(N+1) & - & - & - & 0.81 & 0.81 & 0.79 & 0.82 & 0.83 & 0.83 & 0.68 & 0.36 & 0.26 \\
 SSC(N)  & - & - & - & 0.69 & 0.77 & 0.76 & 0.69 & 0.79 & 0.81 & - & - & -\\ \hline
\end{tabular}
\label{tab:faceclustering}
\end{center}
\end{table*}

Table \ref{tab:faceclustering} shows the accuracy evaluation in the adjusted rand index (ARI) \cite{ami}, AMI, and F1 score of inlier/outlier classification.
DBSCAN, IDC, STSC, and the reference methods work well, whereas the other methods do not.
IDC always earned the highest ARI and AMI scores. 
Note that IDC earned a better score than that of STSC for $K=5$ despite the worse result in the average error of the cluster number estimation. This indicates that IDC finds reasonable clustering results even when it fails to estimate the true cluster numbers.
In F1 score, DBSCAN earned the best scores, and IDCs are the second best for any cluster numbers. Although SC(N+1) is compatible to the scores by DBSCAN, IDCs cannot outperform DBSCAN. This will be due to the errors in the cluster number estimation. 
Note that the F1 scores by IDC decreases with increasing number of inlier clusters in this experimental setting. More precisely, the potential outlier association $|V_{out}|/|V|$ is decreased by the increased number of inliers.
Even under such situation, IDC \hlah{without any parameter settings} achieved the scores close to DBSCAN with parameter tuning.
It is reserved as a future work to investigate IDC's behavior for different numbers of outliers in real data. 


\begin{table*}[t]
\caption{Accuracy of person re-identification in four criteria (average value in 100 trials). }
\begin{center}
\begin{tabular}{c||ccc|ccc|ccc|ccc} \hline 
score & \multicolumn{3}{c|}{Err. in \# of clusters} &\multicolumn{3}{c|}{ARI} & \multicolumn{3}{c}{AMI} & \multicolumn{3}{c}{F1 score}\\ \hline
$N_{in}$ & 4 & 8 & 12 & 4 & 8 & 12 & 4 & 8 & 12 & 4 & 8 & 12\\ \hline
 DBSCAN  & 3.09 & 7.09 & 11.09 & 0.00 & 0.00 & 0.00 & 0.00 & 0.00 & 0.00 & 0.01 & 0.00 & 0.00  \\
 SG  & 2.86 & 6.92 & 10.96 & 0.05 & 0.03 & 0.02  & 0.06  & 0.04 & 0.03 & - & - & - \\ 
 STSC & 11.46 & 9.39 & 7.55 & 0.19 & 0.20 & 0.17 & 0.29 & 0.34 & 0.33 & - & - & -  \\
 Modularity & 3.00 & 7.00 & 11.00 & 0.00 & 0.00 & 0.00 & 0.00 & 0.00 & 0.00 & - & - & -  \\
 SEA & \textbf{2.58} & 6.42 & 10.42 & 0.01 & 0.08 & 0.07 & 0.02 & 0.02 & 0.02 & 0.05 & 0.01 & 0.01 \\
 IDC (ours)& 5.16 & \textbf{5.65} & \textbf{4.01} & \textbf{0.25} & \textbf{0.28}  & \textbf{0.25} & \textbf{0.34} & \textbf{0.43} & \textbf{0.43} & \textbf{0.11} & \textbf{0.06} & \textbf{0.04} \\
 \hline
 SC(N+1) & - & - & - & 0.23 & 0.24 & 0.24 & 0.29 & 0.39 & 0.43 & 0.11 & 0.07 & 0.04 \\
 SSC(N) & - & - & - & 0.09 & 0.11 & 0.12 & 0.15 & 0.24 & 0.30 & - & - & - \\
 
\end{tabular}
\label{tab:personidentification}
\end{center}
\end{table*}

\subsection{Person Re-identification in a Camera Networks}
\label{seq:personidentification}
We synthesized similar dataset with a face clustering senario using Shinpuhkan 2014 dataset \cite{kawanishi2014shinpuhkan}, which contains images of pedestrians observed by a camera network in a real shopping mall.
The camera network comprises 16 non-overlapping camera views, with 24 people.

For the porpose of testing methods in a different feature space, we used AlexNet \cite{alexnet} to extract visual features.
\hlahh{This feature has 4,096 dimensions and non-sparse, and in strong contrast to the 128 dimensional sparse features used in face clustering.
Note that our purpose is not to achieve precise person re-identification but to test clustering methods by pre-fine-tuned model, which is a modern feature representation for general objects, and can be a choice to represent unsupervised visible objects. For readers interested in the state-of-the-art fine-tuned model for person re-identification, please refer} \cite{xiao2016learning}. \hlahh{In the preliminary experiments, we confirmed that DBSCAN with a proper parameter setting can solve this problem.}\footnote{\hlah{By using scikit-learn, DBSCAN with the parmeters of $eps=0.25$ and $min\_samples=2$ earned 0.98 AMI score in a task of clustering 12 people with outliers, where the model is trained in the same manner as the original paper other than the use of only four people from the Shinpuhkan dataset (to keep other 20 as the clustering targets).}}
Note also that an observation of a person in a camera is provided as a tracklet, an in-camera tracking result. 
\hlahh{To construct affinity matrices from tracklet comparison, we used median of affinity values obtained from all frame pairs between the two tracklets.}
%



\par
The inlier/outlier synthesis was conducted for $N_{in}$= 4, 8, and 12. We used ${\cal N}(\mu = 8,\sigma = 1)$ to decide the size of inlier clusters and the number of outliers.
The search range for $\hat{K}$ is set to $K_{min}=3$ and $K_{max}=20$.
\par


Table \ref{tab:personidentification} shows the average error in cluster number estimation and the accuracy evaluation for ARI, AMI, and F1 score. 
Although IDC yielded %
the large average error for $N_{in}=4$, it earned better AMI score than SC(N+1) and any other methods.
This indicates that IDC automatically identified reasonable numbers of clusters to divide the given data.
IDC also earned the best ARI and AMI scores for $N_{in}=8$ and 12. STSC earned the second best scores. In contrast, DBSCAN earned 0.00 scores, which is the same as that earned by clustering by chance.
This indicates that only IDC and STSC could find significant data consensus from deep-learning-based features without supervision.



\section{Conclusion}
In this paper, \hlah{we mathematically observed spectral clustering's properties that are useful for outlier detection and cluster number estimation. Based on these observations, we designed a simple function that evaluates clustering results and experimentally confirmed the effectiveness of the properties.}
Though DBSCAN is an excellent known method, it does not work properly with high-dimensional data.
A graph-based clustering method is expected to work with such data, since it can normalize cluster density not by geometric volume but by the number of edges. Among such methods, spectral clustering is known as a time-tested method that has a rigorous mathematical background.
It is also one of the few clustering methods that work with deep-learning-based features, but there have been no spectral-clustering-based methods for detecting outliers and estimating the number of clusters at the same time.
To overcome this problem, we determined the property of spectral clustering that forms the outlier cluster when inlier clusters are properly isolated with a given $K$.
Based on this property, we developed IDC that evaluates the outlier cluster formation and identifies the number of clusters.

In the experiment, IDC is evaluated in two tasks: face clustering in photo albums and person re-identification with features extracted by a non-fine-tuned model. 
The results show that IDC works with different deep-learning-based features. It estimates the number of clusters even for large cluster numbers and outperforms state-of-the-art methods in many respects under various conditions.


\hlah{
Although the proposed method outperformed the comparative methods, there is room for improvement.
First, for the purpose of outlier detection, there can be more suitable affinity metric, affinity rescaling strategy, and eigendecomposition methods than that used in the experiments.
Second, the evaluation function inspired by isolated pseudo clique is heuristic, and not mathematically optimal.
The searching strategy can also be improved in for computational efficiency.
All these points will be considered for a future work.
}

\section*{Acknowledgement}
This work was supported by CREST, JST.

\bibliographystyle{icml2017}
\bibliography{library}

\begin{thebibliography}{31}
\providecommand{\natexlab}[1]{#1}
\providecommand{\url}[1]{\texttt{#1}}
\expandafter\ifx\csname urlstyle\endcsname\relax
  \providecommand{\doi}[1]{doi: #1}\else
  \providecommand{\doi}{doi: \begingroup \urlstyle{rm}\Url}\fi

\bibitem[Assent(2012)]{assent2012clustering}
Assent, Ira.
\newblock Clustering high dimensional data.
\newblock \emph{Wiley Interdisciplinary Reviews: Data Mining and Knowledge
  Discovery}, 2\penalty0 (4):\penalty0 340--350, 2012.

\bibitem[Bishop(2006)]{Christopher2006}
Bishop, Christopher~M.
\newblock Pattern recognition and machine learning.
\newblock \emph{Company New York}, 16\penalty0 (4):\penalty0 049901, 2006.

\bibitem[Blei \& Jordan(2006)Blei and Jordan]{Blei2006}
Blei, David~M. and Jordan, Michael~I.
\newblock Variational inference for dirichlet process mixtures.
\newblock \emph{Bayesian analysis}, 1\penalty0 (1):\penalty0 pp.121--143, 2006.

\bibitem[Brandes et~al.(2006)Brandes, Delling, Gaertler, G{\"o}rke, Hoefer,
  Nikoloski, and Wagner]{brandes2006maximizing}
Brandes, Ulrik, Delling, Daniel, Gaertler, Marco, G{\"o}rke, Robert, Hoefer,
  Martin, Nikoloski, Zoran, and Wagner, Dorothea.
\newblock Maximizing modularity is hard.
\newblock \emph{arXiv preprint physics/0608255}, 2006.

\bibitem[Ester et~al.(1996)Ester, Kriegel, Sander, Xu,
  et~al.]{ester1996density}
Ester, Martin, Kriegel, Hans-Peter, Sander, J{\"o}rg, Xu, Xiaowei, et~al.
\newblock A density-based algorithm for discovering clusters in large spatial
  databases with noise.
\newblock In \emph{Proceedings of the 2nd International Conference on Knowledge
  Discovery and Data Mining}, pp.\  226--231, 1996.

\bibitem[Florian~Schroff \& Philbin(2015)Florian~Schroff and
  Philbin]{florian2015openface}
Florian~Schroff, Dmitry~Kalenichenko and Philbin, James.
\newblock Facenet: A unified embedding for face recognition and clustering.
\newblock \emph{CVPR}, 2015.

\bibitem[Hai \& Qingsheng(2012)Hai and Qingsheng]{Hai2012}
Hai, Lin and Qingsheng, Zhu.
\newblock A spectral clustering-based dataset structure analysis and outlier
  detection progress.
\newblock \emph{Journal of Computational Information Systems}, 8\penalty0
  (1):\penalty0 115--124, 2012.

\bibitem[Hennig et~al.(2015)Hennig, Meila, Murtagh, and
  Rocci]{hennig2015handbook}
Hennig, Christian, Meila, Marina, Murtagh, Fionn, and Rocci, Roberto.
\newblock \emph{Handbook of cluster analysis}.
\newblock CRC Press, 2015.

\bibitem[Huang et~al.(2007)Huang, Ramesh, Berg, and Learned-Miller]{LFW}
Huang, Gary~B., Ramesh, Manu, Berg, Tamara, and Learned-Miller, Erik.
\newblock Labeled faces in the wild: A database for studying face recognition
  in unconstrained environments.
\newblock Technical Report 07-49, University of Massachusetts, Amherst, October
  2007.

\bibitem[H{\"u}ffner et~al.(2009)H{\"u}ffner, Komusiewicz, Moser, and
  Niedermeier]{huffner2009isolation}
H{\"u}ffner, Falk, Komusiewicz, Christian, Moser, Hannes, and Niedermeier,
  Rolf.
\newblock Isolation concepts for clique enumeration: Comparison and
  computational experiments.
\newblock \emph{Theoretical Computer Science}, 410\penalty0 (52):\penalty0
  5384--5397, 2009.

\bibitem[Ito \& Iwama(2009{\natexlab{a}})Ito and Iwama]{Ito2009}
Ito, Hiro and Iwama, Kazuo.
\newblock Enumeration of isolated cliques and pseudo-cliques.
\newblock \emph{ACM Transactions on Algorithms}, 5\penalty0 (4), October
  2009{\natexlab{a}}.

\bibitem[Ito \& Iwama(2009{\natexlab{b}})Ito and Iwama]{ito2009enumeration}
Ito, Hiro and Iwama, Kazuo.
\newblock Enumeration of isolated cliques and pseudo-cliques.
\newblock \emph{ACM Transactions on Algorithms (TALG)}, 5\penalty0
  (4):\penalty0 40, 2009{\natexlab{b}}.

\bibitem[Ito et~al.(2005)Ito, Iwama, and Osumi]{ito2005linear}
Ito, Hiro, Iwama, Kazuo, and Osumi, Tsuyoshi.
\newblock Linear-time enumeration of isolated cliques.
\newblock In \emph{European Symposium on Algorithms}, pp.\  119--130. Springer,
  2005.

\bibitem[Kawanishi et~al.(2014)Kawanishi, Wu, Mukunoki, and
  Minoh]{kawanishi2014shinpuhkan}
Kawanishi, Yasutomo, Wu, Yang, Mukunoki, Masayuki, and Minoh, Michihiko.
\newblock Shinpuhkan2014: A multi-camera pedestrian dataset for tracking people
  across multiple cameras.
\newblock In \emph{20th Korea-Japan Joint Workshop on Frontiers of Computer
  Vision}, 2014.

\bibitem[Krizhevsky et~al.(2012)Krizhevsky, Sutskever, and Hinton]{alexnet}
Krizhevsky, Alex, Sutskever, Ilya, and Hinton, Geoffrey~E.
\newblock Imagenet classification with deep convolutional neural networks.
\newblock In \emph{Advances in neural information processing systems}, pp.\
  1097--1105, 2012.

\bibitem[Li \& Chen(2015)Li and Chen]{li2015superpixel}
Li, Zhengqin and Chen, Jiansheng.
\newblock Superpixel segmentation using linear spectral clustering.
\newblock In \emph{2015 IEEE Conference on Computer Vision and Pattern
  Recognition (CVPR)}, pp.\  1356--1363. IEEE, 2015.

\bibitem[Li et~al.(2007)Li, Liu, Chen, and Tang]{li2007noise}
Li, Zhenguo, Liu, Jianzhuang, Chen, Shifeng, and Tang, Xiaoou.
\newblock Noise robust spectral clustering.
\newblock In \emph{ICCV2007}, pp.\  1--8, 2007.

\bibitem[Liu et~al.(2013)Liu, Latecki, and Yan]{Hairong2013}
Liu, Hairong, Latecki, Longin~Jan, and Yan, Shuicheng.
\newblock Fast detection of dense subgraphs with iterative shrinking and
  expansion.
\newblock \emph{IEEE Transactions on Pattern Analysis and Machine
  Intelligence}, 35\penalty0 (9), September 2013.

\bibitem[Narayana et~al.(2013)Narayana, Hanson, and
  Learned-Miller]{narayana2013coherent}
Narayana, Manjunath, Hanson, Allen, and Learned-Miller, Erik.
\newblock Coherent motion segmentation in moving camera videos using optical
  flow orientations.
\newblock In \emph{Proceedings of the IEEE International Conference on Computer
  Vision}, pp.\  1577--1584, 2013.

\bibitem[Newman(2006)]{Newman2006}
Newman, M.E.
\newblock Modularity and community structure in networks.
\newblock \emph{Proceeding of the National Academy of Sciences}, 103\penalty0
  (23):\penalty0 8577--8582, 2006.

\bibitem[Ng et~al.(2002)Ng, Jordan, and Weiss]{Andrew2002}
Ng, Andrew~Y., Jordan, Michael~I., and Weiss, Yair.
\newblock On spectral clustering: Analysis and an algorithm.
\newblock \emph{In Neural Information Processing Systems}, pp.\  849--856,
  2002.

\bibitem[Parsons et~al.(2004)Parsons, Haque, and Liu]{parsons2004subspace}
Parsons, Lance, Haque, Ehtesham, and Liu, Huan.
\newblock Subspace clustering for high dimensional data: a review.
\newblock \emph{ACM SIGKDD Explorations Newsletter}, 6\penalty0 (1):\penalty0
  90--105, 2004.

\bibitem[Pelleg \& Moore(2000)Pelleg and Moore]{Pelleg2000}
Pelleg, Dan and Moore, Andrew.
\newblock X-means: Extending k-means with efficient estimation of the number of
  clusters.
\newblock \emph{ICML}, 1, 2000.

\bibitem[Shi \& Malik(2000)Shi and Malik]{Jianbo2000}
Shi, Jianbo and Malik, Jitendra.
\newblock Normalized cuts and image segmentation.
\newblock \emph{IEEE Transactions on Pattern Analysis and Machine
  Intelligence}, 22\penalty0 (8), August 2000.

\bibitem[Vinh et~al.(2009)Vinh, Epps, and Bailey]{ami}
Vinh, Nguyen~Xuan, Epps, Julien, and Bailey, James.
\newblock Information theoretic measures for clusterings comparison: is a
  correction for chance necessary?
\newblock In \emph{Proceedings of the 26th Annual International Conference on
  Machine Learning}, pp.\  1073--1080. ACM, 2009.

\bibitem[Wang et~al.(2015)Wang, Gu, and Chen]{wang2015clustering}
Wang, Shulin, Gu, Jinchao, and Chen, Fang.
\newblock Clustering high-dimensional data via spectral clustering using
  collaborative representation coefficients.
\newblock In \emph{International Conference on Intelligent Computing}, pp.\
  248--258. Springer, 2015.

\bibitem[Wang \& Davidson(2010)Wang and Davidson]{Wang2010}
Wang, Xiang and Davidson, Ian.
\newblock Active spectral clustering.
\newblock \emph{In ICDM, IEEE}, pp.\  pp.561--568, 2010.

\bibitem[Wu et~al.(2014)Wu, Feng, and Zhou]{wu2014spectral}
Wu, Sen, Feng, Xiaodong, and Zhou, Wenjun.
\newblock Spectral clustering of high-dimensional data exploiting sparse
  representation vectors.
\newblock \emph{Neurocomputing}, 135:\penalty0 229--239, 2014.

\bibitem[Xiao et~al.(2016)Xiao, Li, Ouyang, and Wang]{xiao2016learning}
Xiao, Tong, Li, Hongsheng, Ouyang, Wanli, and Wang, Xiaogang.
\newblock Learning deep feature representations with domain guided dropout for
  person re-identification.
\newblock In \emph{2016 IEEE Conference on Computer Vision and Pattern
  Recognition (CVPR)}, 2016.

\bibitem[Yu \& Shi(2003)Yu and Shi]{Stella2003}
Yu, Stella~X. and Shi, Jianbo.
\newblock Multiclass spectral clustering.
\newblock \emph{International Conference on Computer Vision}, pp.\  11--17,
  October 2003.

\bibitem[Zelnik-Manor \& Perona(2004)Zelnik-Manor and Perona]{zelnik2005self}
Zelnik-Manor, Lihi and Perona, Pietro.
\newblock Self-tuning spectral clustering.
\newblock In \emph{Advances in Neural Information Processing Systems 17 (NIPS
  2004)}. MIT Press, 2004.

\end{thebibliography}

\end{document}